\documentclass[10pt,twocolumn,letterpaper]{article}

\usepackage[pagenumbers]{cvpr} % To force page numbers, e.g. for an arXiv version

\usepackage[utf8]{inputenc}

\usepackage{makecell}
\usepackage{multirow}
\usepackage{colortbl}
\usepackage{textcomp} 
\usepackage{indentfirst}
\usepackage{pifont}

\usepackage{algorithm}
\usepackage{algpseudocode}
\usepackage{amsmath}

\definecolor{nbarrier}{RGB}{255, 120, 50}
\definecolor{nbicycle}{RGB}{255, 192, 203}
\definecolor{nbus}{RGB}{255, 255, 0}
\definecolor{ncar}{RGB}{0, 150, 245}
\definecolor{nconstruct}{RGB}{0, 255, 255}
\definecolor{nmotor}{RGB}{200, 180, 0}
\definecolor{npedestrian}{RGB}{255, 0, 0}
\definecolor{ntraffic}{RGB}{255, 240, 150}
\definecolor{ntrailer}{RGB}{135, 60, 0}
\definecolor{ntruck}{RGB}{160, 32, 240}
\definecolor{ndriveable}{RGB}{255, 0, 255}
\definecolor{nother}{RGB}{139, 137, 137}
\definecolor{nsidewalk}{RGB}{75, 0, 75}
\definecolor{nterrain}{RGB}{150, 240, 80}
\definecolor{nmanmade}{RGB}{213, 213, 213}
\definecolor{nvegetation}{RGB}{0, 175, 0}

\definecolor{sub}{HTML}{E9E7EF}

\definecolor{nvcolor}{RGB}{119,185,0}
\definecolor{roadcolor}{RGB}{234,51,246}
\definecolor{sidewalkcolor}{RGB}{68,8,72}
\definecolor{parkingcolor}{RGB}{241,156,249}
\definecolor{othergroundcolor}{RGB}{160,32,76}
\definecolor{buildingcolor}{RGB}{246,202,69}
\definecolor{carcolor}{RGB}{111,149,238}
\definecolor{truckcolor}{RGB}{74,32,172}
\definecolor{bicyclecolor}{RGB}{136,227,242}
\definecolor{motorcyclecolor}{RGB}{37,59,146}
\definecolor{othervehiclecolor}{RGB}{96,81,242}
\definecolor{vegetationcolor}{RGB}{79, 173, 50}
\definecolor{trunkcolor}{RGB}{126, 65, 22}
\definecolor{terraincolor}{RGB}{171, 238, 105}
\definecolor{personcolor}{RGB}{234, 60, 49}
\definecolor{bicyclistcolor}{RGB}{234, 66, 195}
\definecolor{motorcyclistcolor}{RGB}{138, 42, 90}
\definecolor{fencecolor}{RGB}{238, 128, 69}
\definecolor{polecolor}{RGB}{252, 241, 161}
\definecolor{trafficsigncolor}{RGB}{233, 51, 35}
\definecolor{other-struct.color}{RGB}{255, 150, 0}
\definecolor{other-objectcolor}{RGB}{50, 255, 255}
\definecolor{lane-markingcolor}{RGB}{150, 255, 170}
\definecolor{color1}{RGB}{176, 36, 24}
\definecolor{color2}{RGB}{0, 176, 80}
\definecolor{color3}{RGB}{0, 0, 200}

\newcommand{\myparagraph}[1]{\noindent{\bf #1}}

\definecolor{cvprblue}{rgb}{0.21,0.49,0.74}
\usepackage[pagebackref,breaklinks,colorlinks,citecolor=cvprblue]{hyperref}

\usepackage{adjustbox}
\usepackage{multirow}
\usepackage{colortbl}
\usepackage{xspace}
\usepackage[capitalize]{cleveref}
\usepackage{multirow}
\usepackage{float}
\usepackage{booktabs}
\usepackage[table, HTML]{xcolor}
\usepackage{bbding} 
\usepackage{pifont} 

\usepackage{amssymb}  

\def\paperID{***} % *** Enter the Paper ID here
\def\confName{CVPR}
\def\confYear{2026}

\title{DLWM: Dual Latent World Models enable Holistic Gaussian-centric \\ Pre-training in Autonomous Driving}
\author{
\begin{tabular}[t]{@{}c@{}}
Yiyao Zhu$^{1*}$,
Ying Xue$^{2*}$,
Haiming Zhang$^2$,
Guangfeng Jiang$^3$,
Wending Zhou$^4$,
Xu Yan$^{4\textsuperscript{\ding{41}}}$\\
Jiantao Gao$^{4}$,
Yingjie Cai$^{4}$,
Bingbing Liu$^{4}$,
Zhen Li$^{2\textsuperscript{\ding{41}}}$,
Shaojie Shen$^{1}$
\end{tabular}\\[1ex]
\begin{tabular}[t]{@{}c@{}}
$^1$HKUST~~
$^2$CUHK-SZ~~
$^3$USTC~~
$^4$Huawei Foundation Model Department
\end{tabular}\\[0.5ex]
{\tt\small yzhucp@connect.ust.hk, 
yanxu44@huawei.com, lizhen@cuhk.edu.cn}
}

\makeatletter
\newcommand\blfootnote[1]{%
  \begingroup
  \renewcommand\@makefnmark{}%
  \footnotetext{#1}%
  \endgroup
}
\makeatother

\begin{document}
\maketitle
\begin{abstract}
Vision-based autonomous driving has gained much attention due to its low costs and excellent performance.
Compared with dense BEV (Bird’s Eye View) or sparse query models, Gaussian-centric method is a comprehensive yet sparse representation by describing scene with 3D semantic Gaussians.
In this paper, we introduce \textbf{DLWM}, a novel paradigm with \textbf{D}ual \textbf{L}atent \textbf{W}orld \textbf{M}odels specifically designed to enable holistic gaussian-centric pre-training in autonomous driving using two stages. In the first stage, DLWM predicts 3D Gaussians from queries by self-supervised reconstructing multi-view semantic and depth images. Equipped with fine-grained contextual features, in the second stage, two latent world models are trained separately for temporal feature learning, including Gaussian-flow-guided latent prediction for downstream occupancy perception and forecasting tasks, and ego-planning-guided latent prediction for motion planning.
Extensive experiments in SurroundOcc and nuScenes benchmarks demonstrate that DLWM shows significant performance gains across Gaussian-centric 3D occupancy perception, 4D occupancy forecasting and motion planning tasks.

\end{abstract}    
\blfootnote{* Equal contribution.}
\blfootnote{$\textsuperscript{\ding{41}}$ Corresponding authors.}
\section{Introduction}
\label{sec:intro}
% Voxel-based methods
% BEV-based methods
% Sparse query-based methods
% Gaussian-centric representation
% comprehensive yet sparse representation
Recently, vision-based autonomous driving systems~\cite{yan20222dpass, yan2024forging, hu2025vision} have matured into a dominant paradigm that offers a cost-effective and scalable alternative to multi-sensor fusion approaches. This system leverages advanced deep learning and is compatible with multi-task heads, maintaining accurate scene understanding and safe motion planning~\cite{jiang2023vad, jia2023driveadapter}. 
A foundational challenge in achieving robust autonomy is developing a scene representation that is simultaneously expressive, efficient, and temporally coherent for perception, forecasting, and planning tasks. 

Early approaches primarily relied on dense or coarse representations: Voxel-based~\cite{zhou2018voxelnet} methods use 3D voxel grids to represent the surroundings, offering detailed geometric information at the cost of computational overhead. BEV-based methods~\cite{li2022bevformer, bevdet} squeeze multi-view features into a 2D plane; follow-up sparse-query~\cite{lin2022sparse4d, liu2023sparsebev} approaches replace the grid with a handful of sparse queries (e.g., instance boxes, map elements). Despite their relative efficiency, these approaches either sacrifice vertical detail and dense geometry or leave the decision module with only coarse scene knowledge. To overcome these limitations, the research moves to Gaussian-centric representation~\cite{huang2024gaussianformer, zheng2024gaussianad}. A set of 3D semantic Gaussians provides a comprehensive yet sparse representation, offering an optimal balance of detail and efficiency.

\begin{figure}[t]
  \centering
  \includegraphics[width=1\columnwidth]{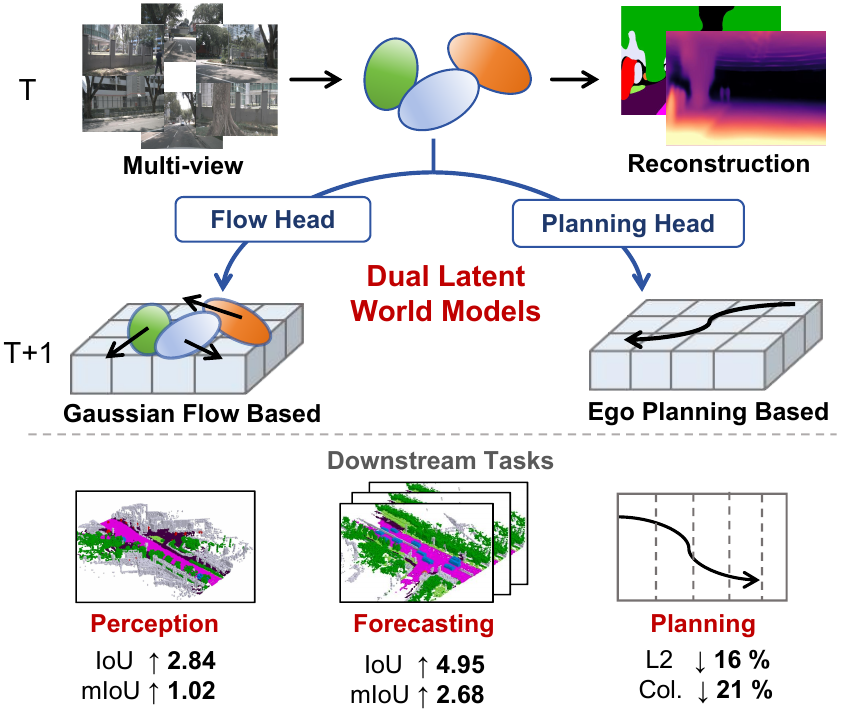}
  \caption{\textbf{Illustration of our DLWM for pre-training and performance improvements for downstream tasks}. }
  \label{fig:fig1}
\end{figure}

% Self-supervised pre-training for gaussian-centric model
Although Gaussian-centric representations have demonstrated significant potential, their reliance on extensive manual annotations hinders scalable deployment. The recent pre-training paradigms utilizing unlabeled data offer a promising solution. For instance, self-supervised approaches like Masked Autoencoders (MAE)~\cite{min2023occupancy} apply contrastive learning for pre-training, but fail to explicitly learn the 3D geometric structures due to rely on coarse supervision. Recently, to learn a complete geometric representation, rendering-based methods such as UniPAD~\cite{yang2024unipad} and ViDAR~\cite{yang2023vidar} utilize LiDAR depth to supervise voxel rendering. In contrast, recent GaussianFlowOcc~\cite{boeder2025gaussianflowocc} and SQS~\cite{zhang2025sqs} show that 3D Gaussians alone can be learned from unlabeled videos through differentiable RGB/depth rendering. Still, a comprehensive self-supervised pre-training strategy tailored for the full lifecycle of Gaussian-centric models remains largely unexplored. 

% Latent World Model introduce
Building upon the necessity of robust feature learning, temporal prediction becomes the next challenge for high-level scene evolution. Latent world model~\cite{li2024enhancing} has emerged as a key approach to unsupervised temporal modeling. It bypasses explicit image or occupancy generation and forecasts future dynamics directly in a compact latent space. Currently, the latent world model has been developed for motion planning~\cite{sun2025echo}, but it has rarely been explored for other critical tasks such as perception~\cite{huang2024gaussianformer} and forecasting~\cite{zhang2024efficient, zhu2023biff}, let alone being integrated with Gaussian-centric models.
% Currently, the Latent World Model has been developed for motion planning, but it has rarely been explored for other critical tasks such as perception and forecasting, let alone Gaussian-centric models. 

% 对于Latent 表征选择，通常来说，鉴于Gaussian query 在当前帧和未来帧各自独立初始化，缺少一一对应关系。Given the permutation-equivalance of Gaussian queries 导致直接监督2帧之间 Gaussian Query 特征不可能。 
% Fortunately, 3DGS 具有任意视角渲染能力，BEV rasterization from sparse Gaussian queries 
% grid representation for 帧间区域对应 保留完整信息, 因此我们选择BEV特征作为Latent 表征。
This integration, however, presents a fundamental technical challenge regarding the selection of latent representations. Since Gaussian queries are initialized independently for the current frame and future frames, they lack a one-to-one correspondence~\cite{gan2025gaussianocc}. Consequently, the permutation equivalence of Gaussian queries renders direct supervision of Gaussian query features between two frames impossible~\cite{huang2024gaussianformer}. Fortunately, 3D Gaussian Splatting (3DGS) possesses arbitrary view rendering capabilities. BEV rasterization~\cite{lu2025toward} derived from sparse Gaussian queries, as a dense grid representation, preserves height information by vertical stacking~\cite{lu2025toward} and allows for clear inter-frame regional correspondence. Thus, we select BEV features as the most suitable latent representation for temporal supervision.
% Regarding the selection of latent representations, typically, given that Gaussian queries are initialized independently for the current frame and future frames, they lack a one-to-one correspondence. Due to the permutation equivalence of Gaussian queries, direct supervision of Gaussian query features between two frames is rendered impossible.
% Fortunately, 3D Gaussian Splatting (3DGS) possesses arbitrary view rendering capabilities. Bird's-Eye View (BEV) rasterization derived from sparse Gaussian queries, as a grid representation for inter-frame regional correspondence, preserves complete information. Thus, we select BEV features as the latent representation.

% dual latent world model
To bridge this gap and fully leverage the benefits of both Gaussian-centric representation and latent world models, we introduce \textbf{DLWM}, a novel holistic pre-training paradigm with \textbf{D}ual \textbf{L}atent \textbf{W}orld \textbf{M}odels as shown in Fig. \ref{fig:fig1}. DLWM utilizes a two-stage approach to unify spatiotemporal Gaussian representation learning, improving all downstream tasks without pre-training (i.e., +1.02 mIoU on occupancy perception, +2.68 mIoU on occupancy forecasting, and -16\% L2 error on motion planning as Fig. \ref{fig:fig1}). Specifically, in the first stage, we reconstruct semantic and depth maps for learning Gaussian context. With the pre-trained weights, in the second stage, dual latent world models are utilized separately for pre-training.  
The first one guided by Gaussian flow is specifically designed for downstream 3D occupancy perception and 4D occupancy forecasting tasks. Another latent world model based on predicted ego trajectory is used for improving motion planning. 

To summarize, we list the contributions of this paper as follows:

\begin{itemize}
    \item We propose \textbf{DLWM}, a self-supervised paradigm for 
    holistic gaussian-centric pre-training, which includes the unified first stage for learning gaussian-centric geometric and semantic representation, followed by training dual latent world model separately in the second stage.
    
    \item We introduce a latent world model guided by gaussian flow and ego motion alignment to learn spatiotemporal gaussian feature representation, specifically designed for downstream occupancy perception and forecasting tasks.
    
    \item We design another latent world model guided by the current Gaussian latent and predicted ego trajectory, jointly improving temporal Gaussian-centric representation and ego trajectory planning.
    
    \item DLWM significantly enhances performance in gaussian-centric occupancy perception, forecasting, and planning tasks, achieving state-of-the-art results on SurroundOcc and nuScenes benchmarks.
\end{itemize}
\section{Related Work}
\label{sec:related}

\subsection{World Models in Autonomous Driving}
World models for autonomous driving are primarily categorized based on their scene representation: 2D image-based versus 3D geometry-based approaches. 
2D image-based world models focus on visual generation. Early efforts like GAIA-1~\cite{hu2023gaia} generated single-view videos. Later research, such as DriveDreamer~\cite{wang2024drivedreamer} and Drive-WM~\cite{wang2024driving}, further enhanced multi-view consistency and temporal coherence. 
3D geometry-based world models focus on spatial dynamics using 4D occupancy forecasting. OccWorld~\cite{zheng2024occworld} and DriveWorld~\cite{min2024driveworld} set baselines by predicting future scenes and ego states. RenderWorld~\cite{yan2024renderworld} designs a more effective VAE for occupancy reconstruction. GaussianWorld~\cite{zuo2024gaussianworld} innovates by modeling scene evolution in a 3D Gaussian space, enhancing perception with scene priors. WPT~\cite{jiang2025wpt} enables policy distillation under the world model.

More recently, latent-based world models~\cite{li2024enhancing, zheng2025world4drive, sun2025echo} take a different approach by bypassing explicit image or occupancy forecasting, predicting future dynamics directly in a compressed latent space. 
%
% The self-supervised LAtent World model (LAW)~\cite{li2024enhancing} enables end-to-end driving by predicting future scene features. World4Drive~\cite{zheng2025world4drive} uses vision foundation models to generate multi-modal planning trajectories without perception supervision. To ensure scene coherence, Echo-Planning~\cite{sun2025echo} proposes a self-correcting framework using a closed-loop Current $\to$ Future $\to$ Current (CFC) cycle for bidirectional self-supervision on latent BEV features. 
%
To the best of our knowledge, the utilization of latent world model for perception and forecasting tasks remain largely unexplored.

\begin{figure*}[t]
  \centering
  \includegraphics[width=2\columnwidth]{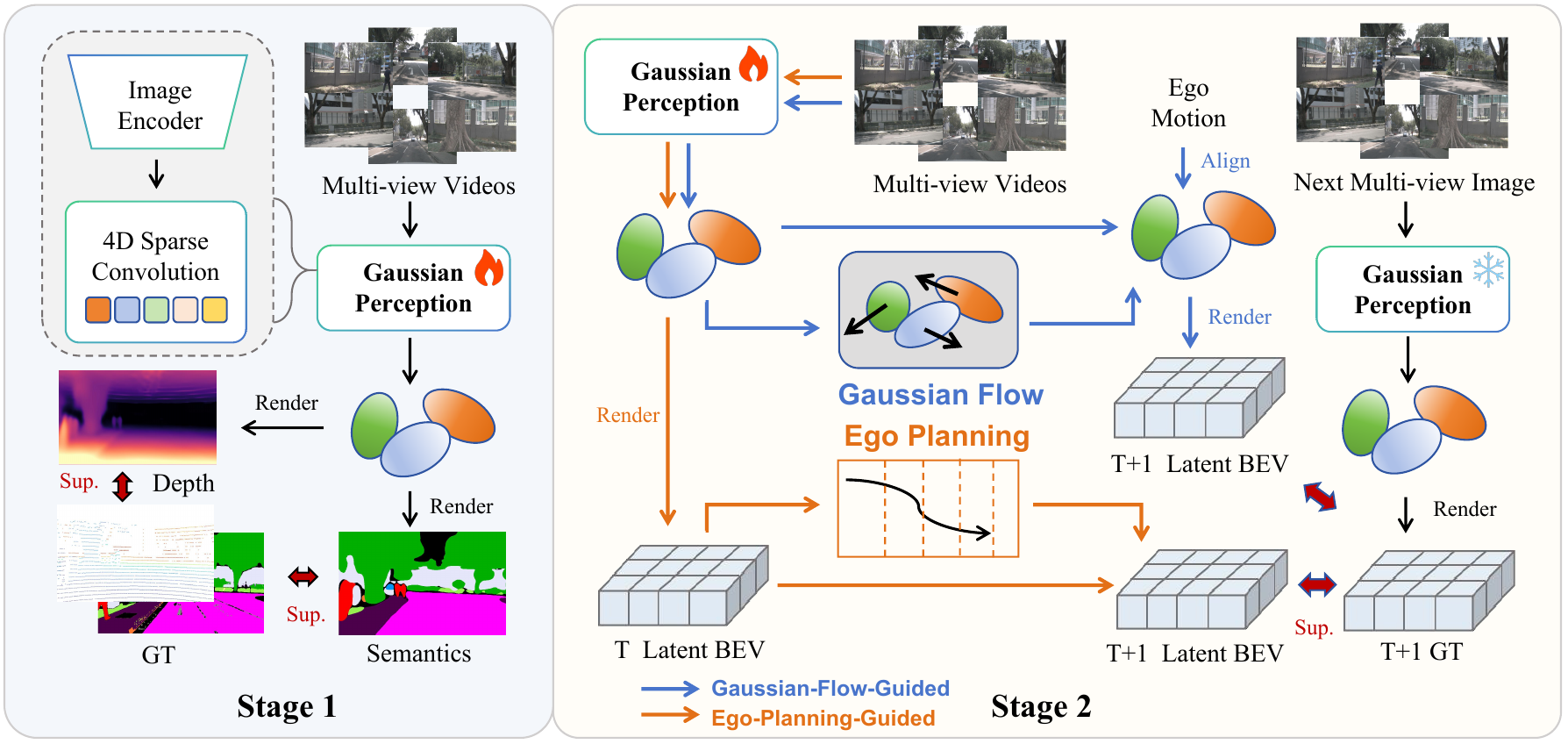}
  \caption{\textbf{Overall pipeline of DLWM.} Stage 1 (Sec. \ref{sec:3.1}) focuses on learning robust 3D Gaussian scene representations from multi-view videos using self-supervised reconstruction on depth and semantic maps. Stage 2 introduces dual latent world models. a. Gaussian-flow-guided model (Sec. \ref{sec:3.2}) explicitly predicts 3D Gaussian flow, propagating the current Gaussian states to the future frame for latent prediction. b. Ego-planning-guided model (Sec. \ref{sec:3.3}) conditions the future scene forecasting on the predicted ego trajectory. All predicted latents are supervised by the perceived features from next multi-view image using a frozen Gaussian perception module.}
  \label{fig:fig2}
\end{figure*}

\subsection{Pre-training in Autonomous Driving}
Pre-training has been a pivotal strategy in autonomous driving, enabling models to learn robust representations from extensive datasets. Particularly, rendering-based pre-training enforces the model to learn physical priors. UniPAD~\cite{yang2024unipad} and MIM4D~\cite{zou2025mim4d} use masked multi-view images and volume rendering for detailed geometric learning. ViDAR~\cite{yang2023vidar} predicts future point clouds from past images, while VisionPAD~\cite{zhang2025visionpad} uses image supervision to learn geometry and motion. More recently, SQS~\cite{zhang2025sqs} introduces a novel pre-training approach tailored for sparse perception models. While various pre-training strategies have improved the foundation of vision-based systems, self-supervised pre-training tailored for full life-cycle Gaussian-centric tasks has yet to be addressed.

\subsection{3D Gaussian Splatting in Autonomous Driving}
3D Gaussian Splatting (3DGS) further advances explicit scene representation in autonomous driving. The explicit and parameter-efficient nature of 3DGS has been effectively utilized for high-fidelity driving scene reconstruction~\cite{zhou2024drivinggaussian, wei2025omni}. More recently, models like GaussianAD~\cite{zheng2024gaussianad} and Gaussianformer~\cite{huang2024gaussianformer} adopted 3DGS for geometric reasoning in tasks like occupancy prediction. Beyond direct perception, 3DGS is increasingly used as a self-supervision representation. GaussianFlowOcc~\cite{boeder2025gaussianflowocc} predicts temporal flow of each Gaussian, effectively learning 4D movement. GaussTR~\cite{jiang2025gausstr} enhances self-supervised understanding by aligning Gaussian representations with 2D models. SQS~\cite{zhang2025sqs} enhances sparse perception models by predicting 3D Gaussians for multi-view reconstruction.

\section{Methodology}

We propose DLWM, a two-stage paradigm for holistic Gaussian-centric pre-training as shown in Fig. \ref{fig:fig2}. In Sec. \ref{sec:3.1}, we present the first stage, which focuses on Gaussian representation learning through the reconstruction of multi-view depth and semantic images. Building upon the pre-trained Gaussian perception module from the first stage, Sec. \ref{sec:3.2} describes a latent world model guided by Gaussian flow pre-trained for downstream tasks such as occupancy perception and forecasting. Subsequently, Sec. \ref{sec:3.3} introduces a separate latent world model in the second stage, dedicated to ego-motion planning, which is trained independently while leveraging the pre-trained perceptual representations from the first stage.

% \begin{figure*}[t]
%   \centering
%   \includegraphics[width=2\columnwidth]{figures/fig2.pdf}
%   \caption{\textbf{Overall pipeline of DLWM.} Stage 1 (Sec. \ref{sec:3.1}) focuses on learning robust 3D Gaussian scene representations from multi-view videos using self-supervised reconstruction on depth and semantic maps. Stage 2 introduces dual latent world models. a. Gaussian-flow-guided model (Sec. \ref{sec:3.2}) explicitly predicts 3D Gaussian flow, propagating the current Gaussian states to the future frame for latent prediction. b. Ego-planning-guided model (Sec. \ref{sec:3.3}) conditions the future scene feature on the predicted ego trajectory. All predicted latent are supervised by the perceived feature from next multi-view image using a frozen Gaussian perception module.}
%   \label{fig:fig2}
% \end{figure*}

\subsection{Gaussian Representation Learning}
\label{sec:3.1}

% During the first stage for pre-training, the 
% In this section, we introduce the preliminaries of 3D Gaussian Splatting (3D-GS)~\cite{kerbl20233d}, 

\myparagraph{Preliminaries.}
3D Gaussian Splatting (3DGS)~\cite{kerbl20233d} represents a scene as a collection of $K$ 3D Gaussian ellipsoids. Each primitive $g_k$ is specified by a three-dimensional center \(\mathbf{\mu}_k \in \mathbb{R}^3\), a covariance \(\mathbf{\Sigma}_k\), an opacity \(\alpha_k \in [0,1]\), and a set of spherical harmonic (SH) coefficients \(c_k \in \mathbb{R}^k\).
Therefore, the Gaussian distribution at point $x$ is represented as:
\begin{equation}
    G(x) = \mathrm{exp}(-\frac{1}{2}(x-\mathbf{\mu})^T\mathbf{\Sigma}^{-1}(x-\mathbf{\mu})).
\end{equation}
Here, the covariance is parameterized via a scale–rotation factorization $\mathbf{\Sigma} = \mathbf{R S S}^T \mathbf{R}^T$, where $\mathbf{S} \in \mathbb{R}^3_+$ denotes the scaling factors, $\mathbf{R} \in SO(3)$ is the rotation matrix, parameterized as a quaternion.

% Projected into image coordinates, the transformation is mediated by the view matrix \(\mathbf{W}\) and the Jacobian \(\mathbf{J}\), yielding:
% % \[
% % \boldsymbol{\Sigma}' = \mathbf{J} \, \mathbf{W} \, \boldsymbol{\Sigma} \, \mathbf{W}^\top \mathbf{J}^\top.
% % \]
% \begin{equation}
%     \Sigma' = \mathbf{J W \Sigma W}^T \mathbf{J}^T.
% \end{equation}

When rendering, we aggregate the contributions of the Gaussians in a depth-ordered fashion via an alpha-blending procedure~\cite{mildenhall2020nerf}. 
% So the pixel color for location \(p\) can be computed as
% \begin{equation}
%     \mathbf{C}(p) = \sum_{i=1}^{K} c_i \,\alpha_i \prod_{j=1}^{i-1} (1 - \alpha_j).
% \end{equation}
The rendering pipeline can be readily adapted to facilitate the retrieval of geometric and semantic information, via:
\begin{align}
    \mathbf{D}(p) = \sum_{i=1}^{K} d_i \,\alpha_i \prod_{j=1}^{i-1} (1 - \alpha_j), \\
    \mathbf{S}(p) = \sum_{i=1}^{K} s_i \,\alpha_i \prod_{j=1}^{i-1} (1 - \alpha_j),
\end{align}
where $\mathbf{D}$ and $\mathbf{S}$ represent the accumulated depth and semantic logits, respectively. \(d_i\) denotes the distance from the \(i\)-th Gaussian to the camera. 
%In contrast to volume rendering approaches~\cite{ben2020nerf}, 3DGS employs a more efficient splat-based rasterization scheme that projects three-dimensional Gaussians as two-dimensional image patches.

\myparagraph{Gaussians From Images.}
Existing dense 3D voxel representations suffer from high overhead and poor resource allocation. We address this by adopting a sparse, explicit representation based on 3D semantic Gaussians, following the design of GaussianFormer~\cite{huang2024gaussianformer}. Each Gaussian query is a sparse, learnable vector used to predict its mean, covariance, opacity and semantic logits. 

We introduce the details of the Gaussian perception module in Fig. \ref{fig:fig2}. At each timestep, we first feed multi-view images through an Image Encoder, which consists of a backbone network (e.g., ResNet101) and a Feature Pyramid Network (FPN), to extract multi-level features. A Gaussian encoder then refines the features block-by-block. Each Gaussian block has three parts: 
1) a self-encoding module for Gaussian–to-Gaussian interaction. This interaction captures the relations of Gaussians within the current frame as well as their temporal propagation from the previous frame through a 4D Sparse Convolution module.
2) an image cross-attention module that aggregates current visual information for deformable feature sampling with camera intrinsics and extrinsics, 
and 3) a refinement module that adjusts Gaussian attributes. 
These queries are transformed into a set of fine-grained Gaussians, and propagated to future timesteps in a streaming fashion.

\myparagraph{Reconstruction Loss for the First Stage.} The first stage is trained via self-supervision, compelling the predicted 3D Gaussians to reconstruct depth and semantic maps. We compute the depth loss $\mathcal{L}_d$ only at pixels with valid LiDAR data, which serves as the ground truth for depth. Given the sparsity of LiDAR-based depth supervision, we leverage perceptual pseudo-depth generated by Metric3D~\cite{yin2023metric3d} as a dense supervisory signal for loss item $\mathcal{L}_{pd}$.
The semantic labels are automatically generated using foundation models Grounded SAM~\cite{ren2024grounded}.
The overall loss is a combination of L1 Loss on depth and cross-entropy loss on semantics without any human label: 
\begin{equation}
\label{eq:recloss}
\mathcal{L}_{rec} = \omega_1 \mathcal{L}_{d} + \omega_2 \mathcal{L}_{pd} + \omega_3 \mathcal{L}_{sem},
\end{equation}
where $\omega_1$, $\omega_2$ and $\omega_3$ are set to 1.0, 0.05 and 1.0 respectively.

\subsection{Latent World Model Guided by Gaussian Flow}
\label{sec:3.2}

For the second stage, given the pre-trained gaussian perception weights in the first stage, we design a latent world model guided by Gaussian flow indicated with blue arrows in Fig. \ref{fig:fig2}. In the end, the pre-trained perception model with enhanced spatiotemporal representation is prepared for improving the performance of downstream 3D occupancy perception and 4D occupancy forecasting tasks. 

\myparagraph{Gaussian Flow Prediction.} Our primary task is to efficiently predict the motion of 3D Gaussians within the scene. It begins by feeding the current perceived Gaussian query features into a flow prediction head. This network is tasked with estimating a local, dynamic displacement vector for each Gaussian. The estimated dynamic displacement, termed the Gaussian flow, will be propagated to the next frame via ego motion alignment: 
\begin{equation}
\label{eq:flowpred}
\mu_{k}^{t+1} = \mathbf{T}_{ego}^{t \to t+1} (\mu_{k}^{t} + \mathbf{\Delta} \mu_k^t),
\end{equation}
where $\mu_{k}^{t} \in \mathbb{R}^3$ and $\mu_{k}^{t+1} \in \mathbb{R}^3$ are the mean of each Gaussian in the current ego frame $t$ and next $t+1$ respectively, $\mathbf{\Delta} \mu_k^t$ is the predicted Gaussian flow, and $\mathbf{T}_{ego}^{t \to t+1} \in \mathbb{R}^{4 \times 4}$ is the coordinate transformation from the current frame to the next frame.

Similar to the reconstruction loss Eq. \ref{eq:recloss} in the first stage, the predicted 3D Gaussians are supervised by rendering depth and semantic images for the next frame.

\myparagraph{Future Latent Prediction.} The next step is to convert the predicted 3D scene into a compact latent representation. After moving and transforming current 3D Gaussians to the next frame, we take the resulted 3D Gaussian set as input, and all other attributes maintain consistency with the current frame. Motivated by GaussianLSS~\cite{lu2025toward}, we perform BEV rasterization to project and rasterize feature of each Gaussian query onto a 2D BEV plane, as the predicted latent in the world model.

\myparagraph{Future Latent Supervision.} We utilize a $\mathbf{L_2}$ loss to supervise the propagated and rasterized future BEV latent feature $\hat{B}_{t+1}$ against the corresponding ground-truth BEV feature $B_{t+1}$, which is defined as the BEV reconstruction loss $\mathcal{L}_{bev}$:
\begin{equation}
\mathcal{L}_{bev} = \| \hat{B}_{t+1} - B_{t+1} \|_2.
\label{eq:bevloss}
\end{equation}

The ground-truth BEV feature $B_{t+1}$ is generated by feeding multi-view image at time $t+1$ into the frozen Gaussian perception module, as shown in the rightmost of Fig. \ref{fig:fig2}. The resulting 3D Gaussians are then rendered and rasterized to produce the true future latent $B_{t+1}$. Minimizing $\mathcal{L}_{bev}$ compels the prediction framework to generate a future scene representation that is both geometrically and semantically consistent with the true future state.

\begin{figure}[t]
  \centering
  \includegraphics[width=1\columnwidth]{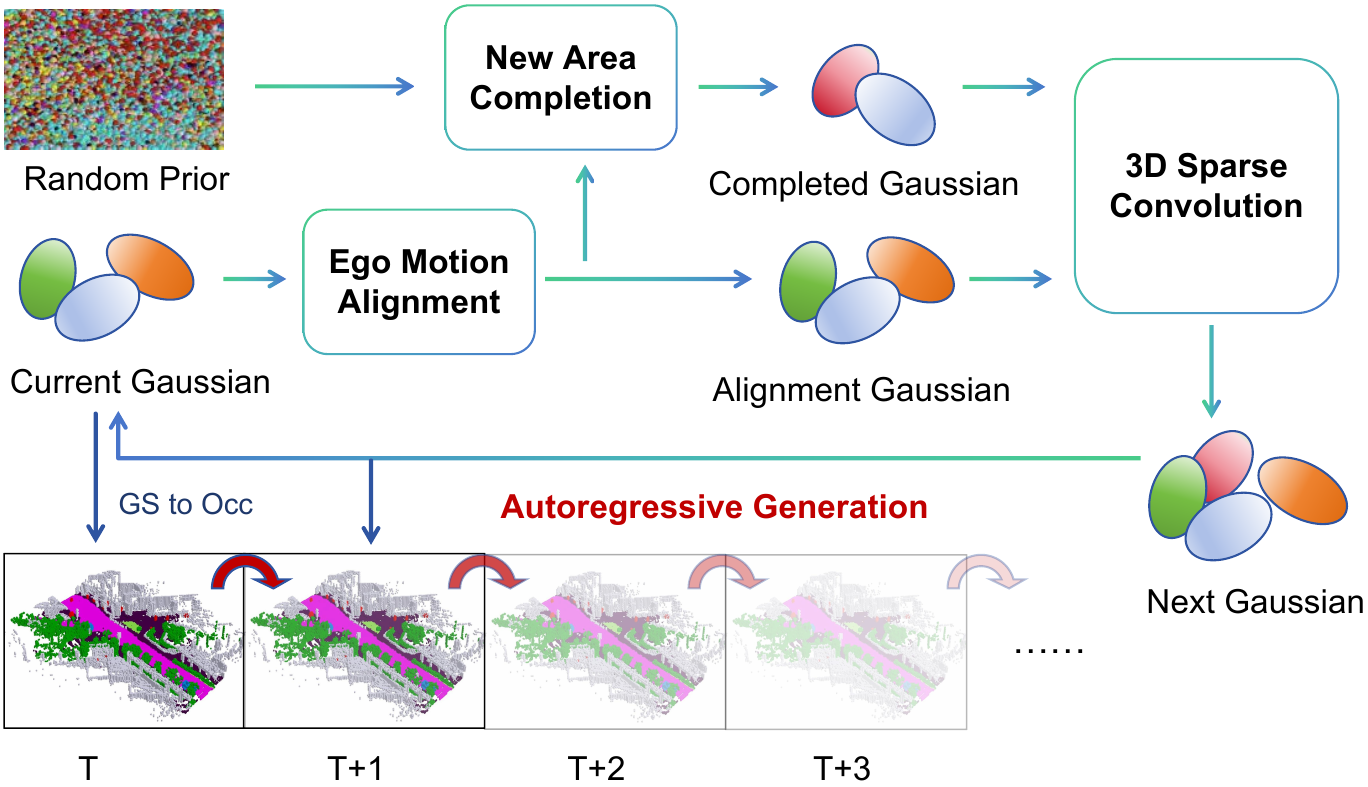}
  \caption{\textbf{4D occupancy forecasting in a stream manner.} We transform current 3D Gaussians to the next frame with ego motion alignment and complete new area with random Gaussians.}
  \label{fig:fig3}
  % \vspace{-0.5em}
\end{figure}

\myparagraph{3D Occupancy Perception.} 
Regarding 3D occupancy perception task, we design a baseline model with the Gaussian perception module in Sec. \ref{sec:3.1} to obtain current 3D Gaussians, followed by Gaussian-to-Occupancy Splatting~\cite{huang2024gaussianformer} for occupancy prediction. 

After the two-stage pre-training, during the fine-tuning stage, we load the pre-trained Gaussian perception module for downstream 3D occupancy perception task. More details can be referred to GaussianFormer~\cite{huang2024gaussianformer} about prediction head and training loss. 

\begin{table*}[t] %
    \small
    \setlength{\tabcolsep}{0.005\linewidth}  
    \renewcommand\arraystretch{1.2}
    \caption{\textbf{3D occupancy perception results on the SurroundOcc-nuScenes validation set \cite{wei2023surroundocc}. } *: We re-evaluate the checkpoints released by GaussianWorld because they repeatedly calculated the metrics for intermediate time-interval frames within each video.
    }
    \centering
    % \vspace{-.5em}
    \resizebox{\textwidth}{!}{
    \begin{tabular}{l|c c | c c c c c c c c c c c c c c c c}
        \toprule
        Method
        & IoU
        & mIoU
        & \rotatebox{90}{\textcolor{nbarrier}{$\blacksquare$} barrier}
        & \rotatebox{90}{\textcolor{nbicycle}{$\blacksquare$} bicycle}
        & \rotatebox{90}{\textcolor{nbus}{$\blacksquare$} bus}
        & \rotatebox{90}{\textcolor{ncar}{$\blacksquare$} car}
        & \rotatebox{90}{\textcolor{nconstruct}{$\blacksquare$} const. veh.}
        & \rotatebox{90}{\textcolor{nmotor}{$\blacksquare$} motorcycle}
        & \rotatebox{90}{\textcolor{npedestrian}{$\blacksquare$} pedestrian}
        & \rotatebox{90}{\textcolor{ntraffic}{$\blacksquare$} traffic cone}
        & \rotatebox{90}{\textcolor{ntrailer}{$\blacksquare$} trailer}
        & \rotatebox{90}{\textcolor{ntruck}{$\blacksquare$} truck}
        & \rotatebox{90}{\textcolor{ndriveable}{$\blacksquare$} drive. suf.}
        & \rotatebox{90}{\textcolor{nother}{$\blacksquare$} other flat}
        & \rotatebox{90}{\textcolor{nsidewalk}{$\blacksquare$} sidewalk}
        & \rotatebox{90}{\textcolor{nterrain}{$\blacksquare$} terrain}
        & \rotatebox{90}{\textcolor{nmanmade}{$\blacksquare$} manmade}
        & \rotatebox{90}{\textcolor{nvegetation}{$\blacksquare$} vegetation}
        \\
        \midrule
        MonoScene~\cite{cao2022monoscene} & 23.96 & 7.31 & 4.03 &	0.35& 8.00& 8.04&	2.90& 0.28& 1.16&	0.67&	4.01& 4.35&	27.72&	5.20& 15.13&	11.29&	9.03&	14.86 \\
        
        Atlas~\cite{murez2020atlas} & 28.66 & 15.00 & 10.64&	5.68&	19.66& 24.94& 8.90&	8.84&	6.47& 3.28&	10.42&	16.21&	34.86&	15.46&	21.89&	20.95&	11.21&	20.54 \\
        
        BEVFormer~\cite{li2022bevformer} & 30.50 & 16.75 & 14.22 &	6.58 & 23.46 & 28.28& 8.66 &10.77& 6.64& 4.05& 11.20&	17.78 & 37.28 & 18.00 & 22.88 & 22.17 & {13.80} & {22.21} \\

        TPVFormer~\cite{huang2023tri}  & {30.86} & 17.10 & 15.96&	 5.31& 23.86	& 27.32 & 9.79 & 8.74 & 7.09 & 5.20& 10.97 & 19.22 & {38.87} & {21.25} & {24.26} & {23.15} & 11.73 & 20.81 \\

        OccFormer~\cite{zhang2023occformer} & {31.39} & {19.03} & {18.65} & {10.41} & {23.92} & {30.29} & {10.31} & {14.19} & {13.59} & {10.13} & {12.49} & {20.77} & {38.78} & 19.79 & 24.19 & 22.21 & {13.48} & {21.35} \\
        
        SurroundOcc~\cite{wei2023surroundocc} & {31.49} & {20.30}  & {20.59} & {11.68} & {28.06} & {30.86} & {10.70} & {15.14} & {14.09} & {12.06} & {14.38} & {22.26} & 37.29 & {23.70} & {24.49} & {22.77} & {14.89} & {21.86} \\

        GaussianFormer~\cite{huang2024gaussianformer} & 29.83 & {19.10} & {19.52} & {11.26} & {26.11} & {29.78} & {10.47} & {13.83} & {12.58} & {8.67} & {12.74} & {21.57} & {39.63} & {23.28} & {24.46} & {22.99} & 9.59 & 19.12\\

        GaussianFormer-2~\cite{huang2024probabilistic} & 31.74 & 20.82 & 21.39 & 13.44 & 28.49 & 30.82 & 10.92 & 15.84 & 13.55 & 10.53 & 14.04 & 22.92 & 40.61 & 24.36 & 26.08 & 24.27 & 13.83 & 21.98 \\

        {GaussianWorld*}~\cite{zuo2024gaussianworld} & 32.77 & 21.79 & 21.61 & 13.30 & 27.28 & 31.21 & 13.89 & 16.91 & 13.28 & 11.77 & 14.82 & 23.66 & 41.91 & 24.31 & 28.35 & 26.32 & 15.67 & 24.54 \\
        
        \midrule
        \textbf{Baseline} & 31.77 & 20.83 & 21.17 & 13.82 & 26.74 & 31.95 & 10.90 & 16.03 & 13.94 & 10.87 & 13.44 & 22.52 & 40.79 & 24.28 & 26.06 & 24.13 & 14.05 & 22.55 \\
        
        \textbf{DLWM  (ours)} & \textbf{34.61} & \textbf{21.85} & 21.71 & 13.56 & 27.65 & 31.22 & 12.50 & 16.93 & 12.35 & 11.42 & 13.96 & 24.12 & 42.27 & 24.94 & 27.39 & 26.36 & 16.61 & 26.64 \\
        
        \bottomrule
    \end{tabular}}
    % \vspace{-.5em}
    \label{tab:nuscenes-results}
\end{table*}

\myparagraph{4D Occupancy Forecasting.} Another downstream task is 4D occupancy forecasting. We design a simple model for stream occupancy forecasting based on Gaussian-centric representation. Motivated by GaussianWorld~\cite{zuo2024gaussianworld}, as illustrated in Fig. \ref{fig:fig3}, the future occupancy is forecast in an autoregressive manner using three key steps, including aligning static scenes, moving dynamic objects and completing of new areas. Concretely, we first obtain the current 3D Gaussians perceived from multi-view videos using Gaussian perception module in Sec. \ref{sec:3.1}. We then align current 3D Gaussians to the next frame based on the ego trajectory. Meanwhile, the newly-observed areas are completed with random Gaussians, facilitating the imagination of these new areas. We employ 3D sparse convolution followed by refinement layers to simultaneously model the current Gaussians and the newly-completed Gaussians. Finally, we utilize all refined Gaussians to predict occupancy in the next frame using
Gaussian-to-Occupancy Splatting~\cite{huang2024gaussianformer}. We forecast all future occupancy in an autoregressive manner, and each frame follows the pipeline mentioned above.

During the fine-tuning stage, we load the pre-trained weights from two-stage (Sec. \ref{sec:3.1} and Sec. \ref{sec:3.2}) pre-training for Gaussian perception module, followed by further training stream 4D occupancy forecasting.

\begin{table*}[t]
\setlength{\tabcolsep}{0.008\linewidth}
\renewcommand\arraystretch{1.2}
\caption{\textbf{4D occupancy forecasting results on the SurroundOcc-nuScenes validation set \cite{wei2023surroundocc}.}
Aux. Sup. represents auxiliary supervision.
Avg. computes the average result of 1s, 2s, and 3s.
}
% \vspace{-.5em}
\small
\centering
\begin{tabular}{l|cc|ccccc|ccccc}
\toprule
\multirow{2}{*}{Method} & \multirow{2}{*}{Input} & \multirow{2}{*}{Aux. Sup.} &
\multicolumn{5}{c|}{mIoU (\%) $\uparrow$} & 
\multicolumn{5}{c}{IoU (\%) $\uparrow$}
\\
&& & 0s & 1s & 2s & 3s & \cellcolor{gray!30}Avg. & 0s & 1s & 2s & 3s & \cellcolor{gray!30}Avg. \\
\midrule
{Copy\&Paste} & 3D-Occ & None & 66.38 & 14.91 & 10.54 & 8.52 & \cellcolor{gray!30}11.33 & 62.29 & 24.47 & 19.77 & 17.31 & \cellcolor{gray!30}20.52 \\
{OccWorld-O}~\cite{zheng2024occworld} & 3D-Occ & None & \textbf{66.38} & \textbf{25.78} & 15.14 & 10.51 & \cellcolor{gray!30}17.14  & \textbf{62.29} & \textbf{34.63} & 25.07 & 20.18 & \cellcolor{gray!30}26.63 \\
\midrule
{OccWorld-T}~\cite{zheng2024occworld} & Camera & Semantic LiDAR & 7.21 & 4.68 &  3.36 & 2.63 & \cellcolor{gray!30}3.56 & 10.66 & 9.32 & 8.23 & 7.47 & \cellcolor{gray!30}8.34 \\
{OccWorld-S}~\cite{zheng2024occworld} & Camera & None & 0.27 & 0.28 & 0.26 & 0.24 & \cellcolor{gray!30}0.26 & 4.32 & 5.05 & 5.01 & 4.95 & \cellcolor{gray!30}5.00 \\

{OccWorld-D}~\cite{zheng2024occworld} & Camera & 3D-Occ & 18.63 & 11.55 & 8.10 & 6.22 & \cellcolor{gray!30}8.62 & 22.88 & 18.90 & 16.26 & 14.43 & \cellcolor{gray!30}16.53 \\

{PreWorld}~\cite{li2025semi} & Camera & 2D Labels \& 3D-Occ & - & 12.27 & 9.24 & 7.15 & \cellcolor{gray!30}9.55 & - & 23.62 & 21.62 & 19.63 & \cellcolor{gray!30}21.62 \\

\midrule
\textbf{Baseline} & Camera & 3D-Occ & 21.12 & 18.18 & 15.11 & 11.97 & \cellcolor{gray!30}15.09 & 32.12 & 29.98 & 25.43 & 21.55 & \cellcolor{gray!30}25.65 \\
\textbf{DLWM (ours)} & Camera & 3D-Occ & 22.21 & 19.66 & \textbf{17.79} & \textbf{15.87} & \cellcolor{gray!30}\textbf{17.77} & 33.47 & 32.63 & \textbf{30.68} & \textbf{28.49} & \cellcolor{gray!30}\textbf{30.60} \\ 

\bottomrule
\end{tabular}

\label{tab:forecast}
% \vspace{-1em}
\end{table*}

\subsection{Latent World Model Guided by Ego Planning}
\label{sec:3.3}

Based on the first stage in Sec. \ref{sec:3.1}, another latent world model in the second stage focuses on predicting future scene features optimized for motion planning. As the orange arrows shown in Fig. \ref{fig:fig2}, this world model is trained independently with the model in Sec. \ref{sec:3.2}. Motivated by LAW~\cite{li2024enhancing} and SSR~\cite{li2024navigation}, we design a forecasting module that uses the predicted trajectory to condition the future latent scene representation, providing motion-aware context for planning.

\myparagraph{Current Latent Construction.} To provide the planning head with a comprehensive, structured, and temporally relevant context, we first transform the current frame's sparse Gaussian features into a dense latent BEV feature via rasterization (as described in Sec. \ref{sec:3.2}). 
To reduce computational load, we extract scene queries $\mathbf{Q}_{\text{scene}}^{t}$ from BEV feature following~\cite{li2024navigation}. 
% This latent feature serves as the input to a transformer decoder, which initializes a set of Scene Queries ($\mathbf{Q}_{\text{scene}}^{t}$) for current scene. 
Next these queries interact via self-attention to capture the spatial relationships and global context of the BEV scene. These contextualized $\mathbf{Q}_{\text{scene}}^{t}$ then serve as the key/value for the waypoint queries $\mathbf{Q}_{\text{wp}}$, coupling the scene perception with the trajectory prediction task.

\myparagraph{Motion Planning.} The objective is to predict the optimal ego trajectory over a future time horizon. The planning head is instantiated by a set of waypoint queries $\mathbf{Q}_{\text{wp}}$. These queries interact with the scene features $\mathbf{Q}_{\text{scene}}^{t}$ via cross-attention to generate the trajectory $\mathbf{\hat{T}}$ through a MLP head for imitation learning with regression: 

\begin{equation}
\label{eq:regloss}
\mathcal{L}_{reg} = \| \mathbf{\hat{T}} - \mathbf{T} \|_1.
\end{equation}

% Latent BEV feature to scene queries

% multi-layer self-attention for scene queries

% cross attn for way point queries

% imitation loss

% Motion aware Layer Normalization 

\myparagraph{Future Latent Prediction.} The core of future prediction is generating the next frame's scene queries $\mathbf{Q}_{\text{scene}}^{t+1}$. We achieve this by conditioning the current scene queries $\mathbf{Q}_{\text{scene}}^{t}$ on the predicted trajectory $\mathbf{\hat{T}}$ using a specialized Motion-Aware Layer Normalization (MLN)~\cite{wang2023exploring} layer; the action $\mathbf{\hat{T}}$ directly modulates $\mathbf{Q}_{\text{scene}}^{t}$'s normalization parameters. 
% This guarantees that the predicted future state is consistent with the ego-vehicle's intended motion, thereby enhancing temporal coherence. 
We predict the next $\mathbf{Q}_{\text{scene}}^{t+1}$ by multi-layer self-attention on motion-aware $\mathbf{Q}_{\text{scene}}^{t}$.
Finally, motivated by~\cite{li2024navigation}, predicted scene queries are fused with the current latent BEV feature $\mathbf{B}_t$ to yield the predicted latent BEV $\hat{B}_{t+1}$ with BEV reconstruction loss $\mathcal{L}_{bev}$ in Eq.~\ref{eq:bevloss}. 

\myparagraph{Total Loss for Motion Planning.} The final losses for motion planning consist of the imitation learning regression loss and the BEV reconstruction loss:

\begin{equation}
\label{eq:planloss}
\mathcal{L}_{plan} = \mathcal{L}_{reg} + \mathcal{L}_{bev}.
\end{equation}

\section{Experiments}

\begin{table*}[t]
\setlength{\tabcolsep}{0.008\linewidth}
\renewcommand\arraystretch{1.15}
\caption{\textbf{Motion Planning Results on the nuScenes validation set. } 
The metrics are computed by the way in VAD~\cite{jiang2023vad}.
$\diamond$: Lidar-based methods.
%Our method (\textbf{Ours}) achieves state-of-the-art performance while requiring \textbf{No Auxiliary Task} outside of the latent world model prediction. 
We do not utilize ego status in the planning module.}
\small
% \vspace{-.5em}
\centering
    \begin{tabular}{l|c|cccc|ccccc}
    \toprule
    \multirow{2}{*}{Method} & 
    \multirow{2}{*}{Aux. Task} & 
    \multicolumn{4}{c}{L2 (m) $\downarrow$} & 
    \multicolumn{4}{c}{Collision Rate (\%) $\downarrow$} \\
    \cmidrule(lr){3-6} \cmidrule(lr){7-10}
    & & 1s & 2s & 3s & \cellcolor{gray!30}Avg. & 1s & 2s & 3s & \cellcolor{gray!30}Avg. \\
    \midrule
    NMP$\diamond$~\cite{zeng2019end} & Det \& Motion & 0.53 & 1.25 & 2.67 & \cellcolor{gray!30}1.48 & \textbf{0.04} & 0.12 & 0.87 & \cellcolor{gray!30}0.34 \\
    FF$\diamond$~\cite{hu2021safe} & FreeSpace & 0.55 & 1.20 & 2.54 & \cellcolor{gray!30}1.43 & 0.06 & 0.17 & 1.07 & \cellcolor{gray!30}0.43 \\
    EO$\diamond$~\cite{khurana2022differentiable} & FreeSpace & 0.67 & 1.36 & 2.78 & \cellcolor{gray!30}1.60 & \textbf{0.04} & 0.09 & 0.88 & \cellcolor{gray!30}0.33 \\
    \midrule
    ST-P3~\cite{hu2022st} & Det \& Map \& Depth & 1.33 & 2.11 & 2.90 & \cellcolor{gray!30}2.11 & 0.23 & 0.62 & 1.27 & \cellcolor{gray!30}0.71 \\
    UniAD~\cite{uniad} & Det\&Track\&Map\&Motion\&Occ & 0.44 & 0.67 & 0.96 & \cellcolor{gray!30}0.69 & \textbf{0.04} & \textbf{0.08} & \textbf{0.23} & \cellcolor{gray!30}\textbf{0.12} \\
    VAD-Tiny~\cite{jiang2023vad} & Det \& Map \& Motion & 0.46 & 0.76 & 1.12 & \cellcolor{gray!30}0.78 & 0.21 & 0.35 & 0.58 & \cellcolor{gray!30}0.38 \\
    VAD-Base~\cite{jiang2023vad} & Det \& Map \& Motion & 0.41 & 0.70 & 1.05 & \cellcolor{gray!30}0.72 & 0.07 & 0.17 & 0.41 & \cellcolor{gray!30}0.22 \\
    PARA-Drive~\cite{weng2024drive} & Det\&Track\&Map\&Motion\&Occ & 0.25 & 0.46 & 0.74 & \cellcolor{gray!30}0.48 & 0.14 & 0.23 & 0.39 & \cellcolor{gray!30}0.25 \\    
    GenAD~\cite{zheng2024genad} & Det \& Map \& Motion & 0.28 & 0.49 & 0.78 & \cellcolor{gray!30}0.52 & 0.08 & 0.14 & 0.34 & \cellcolor{gray!30}0.19 \\
    \midrule
    OccWorld~\cite{zheng2024occworld} & 3D-Occ & 0.39 & 0.73 & 1.18 & \cellcolor{gray!30}0.77 & 0.11 & 0.19 & 0.67 & \cellcolor{gray!30}0.32 \\
    BEV-Planner~\cite{li2024ego} & None & 0.28 & \textbf{0.42} & \textbf{0.68} & \cellcolor{gray!30}\textbf{0.46} & \textbf{0.04} & 0.37 & 1.07 & \cellcolor{gray!30}0.49 \\
    LAW~\cite{li2024enhancing} & None & 0.26 & 0.57 & 1.01 & \cellcolor{gray!30}0.61 & 0.14 & 0.21 & 0.54 & \cellcolor{gray!30}0.30 \\
    \midrule
    \textbf{Baseline} & None & 0.27 & 0.51 & 0.87 & \cellcolor{gray!30}0.55 & 0.11 & 0.17 & 0.46 & \cellcolor{gray!30}0.24 \\
    \textbf{DLWM (ours)} & None & \textbf{0.21} & 0.43 & 0.76 & \cellcolor{gray!30}\textbf{0.46} & 0.11 & 0.14 & 0.32 & \cellcolor{gray!30}0.19 \\
    \bottomrule
    \end{tabular}

\label{tab:comparison}
\vspace{-1em}
\end{table*}

% \begin{figure*}[h]
%   \centering
%   \includegraphics[width=1.7\columnwidth]{figures/vis_occ.pdf}
%   \caption{
%   \textbf{Qualitative comparison of 3D occupancy perception between the model without pre-training (left) and with pre-training (right) on the SurroundOcc-nuScenes validation set.} The pre-trained model predicts more accurate semantic occupancy compared with the model without pre-training (highlighted with black circles).
%   }
%   \label{fig:vis}
% \end{figure*}

\subsection{Datasets}
We evaluate our DLWM using the widely recognized nuScenes~\cite{caesar2020nuscenes} dataset, which comprises 1000 driving sequences. Each sequence contains 20 seconds of video captured by both RGB and LiDAR sensors. The data is recorded at a rate of 20Hz with keyframes at 2Hz. To facilitate evaluation on 3D scene perception and forecasting, we incorporate SurroundOcc~\cite{wei2023surroundocc} with 3D semantic occupancy annotations based on nuScenes. Each voxel in the dataset is labeled with one of 18 categories, consisting of 16 semantic classes, 1 empty class, and 1 unknown class.

\subsection{Evaluation Metrics}
For evaluation, we assess the performance of our model across perception, forecasting, and planning tasks. For perception, the quality of semantic occupancy prediction is evaluated using the mean Intersection-over-Union (mIoU) and Intersection-over-Union (IoU) metrics. For the forecasting task, we adopt the same mIoU and IoU metrics to evaluate the model’s ability to predict future 3D occupancy grids within the next 3 seconds. For the planning task, we evaluate the predicted ego-vehicle trajectories using the L2 error and collision rate, which quantify the accuracy and safety of the generated motion plans.

\subsection{Implementation Details}
We utilize a ResNet101-DCN backbone, initialized from an FCOS3D~\cite{wang2021fcos3d} checkpoint, as the image encoder. Feature extraction is performed using a Feature Pyramid Network (FPN) with downsampled scales of 4, 8, 16, and 32. We set the Gaussian counts to 25,600 and apply four transformer layers to enhance Gaussian attributes for each frame. The model is trained using the AdamW optimizer with a weight decay of 0.01. The learning rate warms up linearly to 4e-4 over the first 500 steps, followed by a cosine decay schedule. Regarding to the pre-training, the first stage is trained for 12 epochs with a batch size of 16. After that,  we further pre-train the perception module using latent world model guided by Gaussian flow with the same training setting. For the fine-tuning, we train our models for 20 epochs for both occupancy perception and forecasting by loading the pre-trained weights, with a batch size of 16. 
Based on the same first stage, another latent world model guided by ego-planning is trained for 20 epochs in the second stage for the final planning results via temporal learning. 

\subsection{Main Results}
We evaluate the effectiveness of DLWM on three challenging downstream tasks: 3D occupancy perception, 4D occupancy forecasting, and motion planning. More visualization results are shown in the supplementary material.
% of 3D occupancy perception can be found in Fig. \ref{fig:vis}. 

\myparagraph{3D Occupancy Perception.} Tab.~\ref{tab:nuscenes-results} compares 3D semantic occupancy results on the nuScenes validation set with SurroundOcc labels. Without pre-training, our baseline model yields 20.83 mIoU and 31.77 IoU. After two-stage pre-training, the model (ours) rises to 21.85 mIoU and 34.61 IoU, achieving SOTA and outperforming the model without pre-training by 1.02 mIoU and 2.84 IoU. The results demonstrate the benefit of our pre-training strategy.

\myparagraph{4D Occupancy Forecasting.} Tab.~\ref{tab:forecast} summarizes the 4D occupancy forecasting results on the nuScenes validation set. We evaluate two variants: Baseline (without pre-training) and DLWM (with two-stage pre-training). The benchmark include Copy\&Paste, OccWorld-O/T/S/D~\cite{zheng2024occworld}.  The Baseline already surpasses all OccWorld variants, yielding 15.09 mIoU and 25.65 IoU averaged over 1–3 s. After pre-training, our approach establishes a new state-of-the-art with an average 17.77 mIoU and 30.60 IoU, outperforming 3D-Occ input method OccWorld-O. The consistent gains confirm the superiority of our pre-trained world model on the 4D occupancy forecasting task.

\myparagraph{Motion Planning.} We evaluate our DLWM on the nuScenes motion planning task, measuring performance via L2 distance and collision rate over a 3-second horizon (Tab.~\ref{tab:comparison}). Our method achieves an average L2 distance of 0.46m, tying for the best score with BEV-Planner and surpassing dedicated world models like LAW (L2: 0.61m). Compared with UniAD with multiple auxillary tasks, our method achieves a better L2 score with comparable collision avoidance, confirming that the designed latent world model is highly effective. Furthermore, our two-stage self-supervised pre-training yields a substantial 0.09m improvement in L2 distance (from 0.55m to 0.46m) over the baseline without pre-training.

% \textbf{Visualization of Renderings.}

\subsection{Ablation Study} In this section, we maintain a consistent data scale to ensure fair comparison. Specifically, we utilize half of the training data during the two-stage pre-training and a quarter for occupancy prediction during the fine-tuning stage, while all training data are used for planning.

\begin{table*}[t]
\caption{\textbf{Ablation study on render supervision and latent world model guided by Gaussian flow.} ``Sup-$D_L$'': sparse depth supervision derived from LiDAR points. ``Sup-$D$'': dense pseudo-depth map supervision. ``Sup-$S$'': semantics supervision. ``Gaussian Flow'': denotes the inclusion of the Stage 2 latent world model guided by Gaussian flow.}
\centering
\footnotesize
\setlength{\tabcolsep}{2mm}
\renewcommand\arraystretch{1.15}
\begin{tabular}{lcccc|cc}
\toprule
\multirow{2}{*}{Exp.} & \multicolumn{4}{c|}{Settings} & \multicolumn{2}{c}{Perception} \\
\cmidrule(lr){2-5} \cmidrule(lr){6-7}
& Sup-$D_L$ & Sup-$D$ & Sup-$S$ & \cellcolor{gray!30}Gaussian Flow & IoU (\%) $\uparrow$ & mIoU (\%) $\uparrow$ \\
\midrule
Exp1 & \checkmark & & & \cellcolor{gray!30} & 18.40 &  29.83 \\
Exp2 & \checkmark & \checkmark &  & \cellcolor{gray!30} & 18.74 & 30.05 \\
Exp3 & \checkmark & \checkmark & \checkmark & \cellcolor{gray!30} & 18.99 & 30.23 \\
Exp4 & \checkmark & \checkmark & \checkmark & \cellcolor{gray!30}\checkmark & \textbf{19.30}  & \textbf{30.56} \\
\bottomrule
\end{tabular}

% \vspace{0.8em}

\label{tab:abl1}
\end{table*}

\begin{table*}[t]
\caption{\textbf{Ablation study on latent world model guided by ego planning.}
Latent BEV: planning with latent BEV rendered from 3D Gaussians.
BEV Forecasting: future latent BEV forecasting integrated into planning.
Stage-1 Pretrain: image reconstruction.
}
\centering
\footnotesize
\setlength{\tabcolsep}{2mm}
\renewcommand\arraystretch{1.15}
\renewcommand\arraystretch{1}
\begin{tabular}{lccccccccccc}
\toprule
\multirow{2}{*}{Exp.} &
\multicolumn{3}{c}{Settings} &
\multicolumn{4}{c}{L2 (m) $\downarrow$} &
\multicolumn{4}{c}{Col. (\%) $\downarrow$} \\
\cmidrule(lr){2-4}\cmidrule(lr){5-8}\cmidrule(lr){9-12}
& Latent BEV & BEV Forecast & Stage-1 Pre-train
& 1s & 2s & 3s & Avg & 1s & 2s & 3s & Avg \\
\midrule
Exp1 &   &   &   & 0.27 & 0.51 & 0.87 & 0.55 & 0.11 & 0.17 & 0.46 & 0.24 \\
Exp2 & \checkmark &   &   & 0.26 & 0.48 & 0.82 & 0.52 & 0.12 & 0.16 & 0.38 & 0.22 \\
Exp3 & \checkmark & \checkmark &   & 0.23 & 0.47 & 0.81 & 0.50 & \textbf{0.10} & \textbf{0.13} & 0.36 & 0.20 \\
Exp4 & \checkmark & \checkmark & \checkmark & \textbf{0.21} & \textbf{0.43} & \textbf{0.76} & \textbf{0.46} & 0.11 & 0.14 & \textbf{0.32} & \textbf{0.19} \\
\bottomrule
\end{tabular}

% \vspace{0.8em}

\label{tab:abl2}
\end{table*}

\begin{table}[h]
\caption{\textbf{Ablation study on data scale.} 3D occupancy perception performance under different pre-training data scales.}
\centering
\footnotesize
\setlength{\tabcolsep}{4mm}
\renewcommand\arraystretch{1.15}
\begin{tabular}{lcc}
\toprule
Dataset fraction & mIoU (\%) $\uparrow$ & IoU (\%) $\uparrow$ \\
\midrule
10\% & 18.24 & 28.59 \\
50\% & 19.30 & 30.56 \\
100\% & \textbf{19.98} & \textbf{31.17} \\
\bottomrule
\end{tabular}
\label{tab:abl3}
\end{table}

\myparagraph{Rendering Objectives.} To investigate the impact of various rendering objectives in Stage 1, 
we conduct an ablation study from Exp1 to Exp3 in Tab.~\ref{tab:abl1}. Starting with the sparse LiDAR depth supervision (Sup-${D_L}$, Exp1, mIoU 18.40), we observe cumulative gains from introducing denser signals (Sup-$D$, Exp2) with mIoU to 18.74, confirming the value of dense geometric supervision. Further incorporating semantic supervision (Sup-$S$, Exp3) yields additional gains, reaching 18.99 mIoU and 30.23 IoU.

\myparagraph{Latent World Model Guided by Gaussian Flow.} We evaluate the contribution of the Gaussian-flow-guided latent world model (Stage 2) for temporal feature learning in Tab.~\ref{tab:abl1} Exp4. By enabling the Gaussian flow module, the final 3D occupancy performance sees a significant boost, achieving a peak 19.30 mIoU and 30.56 IoU. It indicates that Stage 2 Gaussian-flow-guided latent forecasting is crucial for injecting spatio-temporal coherence and predictability into the Gaussian features.

\myparagraph{Latent World Model Guided by Ego Trajectory.} We conduct detailed ablation studies (Tab.~\ref{tab:abl2}) to validate the effectiveness of our motion planning components. Comparing Exp1 (raw Gaussian features) against Exp2 (current latent BEV feature) demonstrates that the rasterized BEV latent representation is superior for the planning head, reducing the average L2 distance from 0.55m to 0.52m and collision rate from 0.24 to 0.21. Integrating future BEV forecasting (Exp3), even without pre-training, significantly improves performance. The most substantial gain is achieved by incorporating the Stage 1 reconstruction pre-training (Exp4, our final method). This pre-training ensures Gaussian features are initialized with rich geometric priors for more accurate scene modeling. As a result, Exp4 achieves the best overall performance, reducing the average L2 distance further to 0.46m and the average collision rate to 0.19. 

\myparagraph{Data Scaling.} We investigate the impact of the data scale of two-stage self-supervised pre-training on the downstream occupancy perception. We vary the proportion of data used for the pre-training between 10\%, 50\%, and 100\% of training data. As shown in Tab.~\ref{tab:abl3}, performance scales positively with data size. Utilizing the full 100\% of the training set yields the best result, achieving a peak mIoU of 19.98 and IoU of 31.17. This confirms that our pre-training leverages unlabeled data to reinforce the geometric and semantic priors of the Gaussians with temporal coherence.

\section{Conclusion}

In this work, we propose DLWM (Dual Latent World Models), a novel two-stage self-supervised pre-training paradigm designed for Gaussian-centric representation for vision-based autonomous driving. DLWM improves sparse query learning and temporal coherence by establishing a two-stage process: Stage 1 focuses on geometric and semantic feature learning via diverse rendering objectives. Stage 2 introduces our dual latent world models including Gaussian-flow-guided and ego-planning-guided latent forecasting. DLWM achieves state-of-the-art results on occupancy perception, forecasting and motion planning tasks, confirming the essential, scalable contribution of our holistic Gaussian-centric pre-training framework.

{
    \small
    \bibliographystyle{ieeenat_fullname}
    \bibliography{main}
}

\clearpage
\setcounter{page}{1}
\renewcommand{\thesection}{\Alph{section}}
\maketitlesupplementary

% \section{Rationale}
% \label{sec:rationale}
% % 
% Having the supplementary compiled together with the main paper means that:
% % 
% \begin{itemize}
% \item The supplementary can back-reference sections of the main paper, for example, we can refer to \cref{sec:intro};
% \item The main paper can forward reference sub-sections within the supplementary explicitly (e.g. referring to a particular experiment); 
% \item When submitted to arXiv, the supplementary will already included at the end of the paper.
% \end{itemize}
% % 
% To split the supplementary pages from the main paper, you can use \href{https://support.apple.com/en-ca/guide/preview/prvw11793/mac#:~:text=Delete%20a%20page%20from%20a,or%20choose%20Edit%20%3E%20Delete).}{Preview (on macOS)}, \href{https://www.adobe.com/acrobat/how-to/delete-pages-from-pdf.html#:~:text=Choose%20%E2%80%9CTools%E2%80%9D%20%3E%20%E2%80%9COrganize,or%20pages%20from%20the%20file.}{Adobe Acrobat} (on all OSs), as well as \href{https://superuser.com/questions/517986/is-it-possible-to-delete-some-pages-of-a-pdf-document}{command line tools}.

% \section{Experiment setup}

% \subsection{Evaluation Metrics}
\section{Evaluation Metrics}

\myparagraph{3D Occupancy Perception.} The primary metrics for 3D occupancy perception are the mean Intersection over Union ($\rm{mIoU}$) and the Intersection over Union ($\rm{IoU}$): 
\begin{equation}
{\rm mIoU} = \frac{1}{|\mathcal{C}'|}\sum_{i\in\mathcal{C}'} \frac{TP_i}{TP_i+FP_i+FN_i},
\end{equation}

\begin{equation}
{\rm IoU}=\frac{TP_{\neq c_0}}{TP_{\neq c_0}+FP_{\neq c_0}+FN_{\neq c_0}},
\end{equation}
where $\mathcal{C}'$ is the set of non-empty classes, $c_0$ is the empty class, and $TP_i$, $FP_i$, and $FN_i$ are the number of true positives, false positives, and false negatives for class $i$, respectively. $TP_{\neq c_0}$, $FP_{\neq c_0}$, and $FN_{\neq c_0}$ are the aggregation of these values across all non-empty classes.

\myparagraph{4D Occupancy Forecasting.}
For the forecasting task, we assess the model's ability to predict scene evolution over a future horizon of 3 seconds, corresponding to 6 frames (at 2Hz). The final metrics are reported as the average of the mIoU and IoU scores computed across these 6 future timestamps:
\begin{equation}
\text{mIoU}_{4D} = \frac{1}{T} \sum_{t=1}^{T} \text{mIoU}_t,
\end{equation}

\begin{equation}
\text{IoU}_{4D} = \frac{1}{T} \sum_{t=1}^{T} \text{IoU}_t,
\end{equation}
where $T$ is the total future time steps, $\text{mIoU}_t$ and $\text{IoU}_t$ denote the metrics calculated at the $t$-th future frame.

\myparagraph{Motion Planning.}
For the motion planning task, we evaluate the quality of the predicted ego-vehicle trajectories using the L2 error and collision rate.  \\
\begin{itemize}
    \item \textbf{L2 Error.} The average Euclidean distance between the predicted waypoints $\hat{T}_t$ and the ground-truth waypoints $T_t$ over the future horizon $N_f$:
    \begin{equation}
        \text{L2} = \frac{1}{N_f} \sum_{t=1}^{N_f} || \hat{T}_t - T_t ||_2
    \end{equation}
    
    \item \textbf{Collision Rate.} A collision occurs if the ego-vehicle's bounding box, at future timestep $N_f$ intersects with the set of all ground-truth obstacle voxels. The final metric is the percentage of scenes where a collision is detected.
\end{itemize}

\section{Implementation Details}

\myparagraph{Details of Gaussian Perception Module.} 
The Gaussian perception module is formulated to construct a compact and high-fidelity 3D scene representation $\mathcal{G}$ parameterized by a set of 3D Gaussians from multi-view image observations. The architecture is composed of an image encoder and a Gaussian transformer decoder.

The image encoder $\mathcal{E}$ takes the multi-view images $\mathcal{I} = \{\mathbf{I}_i \in \mathbb{R}^{3\times H \times W} | i=1,...,N\}$ 
%with camera calibration parameters: intrinsics $\mathcal{K} = \{\mathbf{K}_i\}$ and extrinsics $\mathcal{T} = \{\mathbf{T}_i\}$, 
as input. For each view, a potent backbone (e.g., ResNet-101) first encodes the images into multi-level features $F'$. A Feature Pyramid Network (FPN) then refines $F'$ to generate the final multi-scale image features $F$, which capture both semantic and spatial details.

Central to the Gaussian perception module is a Gaussian transformer decoder. The 3D scene is represented by $K$ distinct, learnable Gaussian queries. Each query $k$ is initialized with a learnable Gaussian anchor $g_k\in \mathbb{R}^{K \times C}$, which constitutes the learnable 3D geometric primitive, and its associated query features $q_k\in \mathbb{R}^{K \times D}$, initially set as zero vectors. $C$ and $D$ are the dimension of Gaussian primitives and query features respectively.
These query features $q_k$ are guided by their corresponding anchors $g_k$ to interact with each other via a self-encoding block, and interact with the image features $F$ via deformable cross-attention to predict the Gaussian attributes.

Specifically, the self-encoding block applies 4D sparse convolution to capture the relations of Gaussians within the current frame, and propagates temporal Gaussian queries from the previous frame in a stream manner.
The deformable cross-attention aggregates the current image information from 2D image to 3D Gaussian queries via camera intrinsics $\mathcal{K}$ and extrinsics $\mathcal{T}$.
%To capture the full spatial extent of the scene with linear complexity, we deploy 4D sparse convolution: each Gaussian anchor is first voxelized according to its 3D position $\mu_k \in \mathbb{R}^3$, after which the convolution operates only on the occupied voxels, yielding efficient, complete 3D aggregation. It then scatters learned offsets around each Gaussian’s mean to obtain several 3D reference points, project them into every image via $\mathcal{K}$ and $\mathcal{T}$, and sample the multi-scale maps. 
Finally, an MLP-based Gaussian head is applied to each Gaussian query, and Gaussian parameters are iteratively refined across decoder layers including position $\mu$, scale $S$, rotation $R$, opacity $\alpha$, and semantics $c$.

\myparagraph{Details of Latent World Model with Gaussian Flow.} To further illustrate the latent world model guided by Gaussian flow, we provide a pseudo code in Alg. \ref{alg:flow}. Given the current Gaussian $G_t$, we aim to predict latent BEV $\hat{B}_{t+1}$ at the next time step. Firstly, we estimate the Gaussian flow $\Delta \mu^t$ by feeding the current Gaussian query features $q_t$ to the flow head. Then in the next frame, we obtain the mean $\mu^{t+1}$ of Gaussians by propagating dynamic flow $\Delta \mu^t$ to the next frame via the ego motion alignment $T_{ego}^{t \rightarrow t+1}$: $T_{ego}^{t \rightarrow t+1}(\mu_k^t + \Delta \mu_k^t)$. To predict the Gaussian set $\hat{G}_{t+1}$, we update the current set $G_t$ with the estimated mean $\mu^{t+1}$ while all the others remain the same. Finally, to obtain the next latent BEV $\hat{B}_{t+1}$, we project the predicted $\hat{G}_{t+1}$ to 2D BEV plane by query feature rasterization~\cite{lu2025toward}.
% \begin{algorithm}[t]
% \caption{Latent World Model with Gaussian Flow}
% \label{alg:gaussian_flow}
% \begin{algorithmic}[1]
% \Require 
%     $G_t = \{\mu_k^t, \Sigma_k^t, \alpha_k^t, c_k^t, q_k^t\}$, 
%     $T_{ego}^{t \rightarrow t+1}$
% \Ensure $\hat{B}_{t+1}$.

% \State \textbf{1. Flow Prediction:}
% \State $\Delta \mu^t \leftarrow \text{FlowHead}(G_t)$ \Comment{Predict dynamic displacement}

% \State \textbf{2. Ego Motion Alignment (Eq. 5):}
% \For{each Gaussian $k$}
%     \State $\mu_k^{t+1} \leftarrow T_{ego}^{t \rightarrow t+1}(\mu_k^t + \Delta \mu_k^t)$ \Comment{Apply flow and ego-motion}
% \EndFor
% \State Update Gaussian set: $\hat{G}_{t+1} \leftarrow \{\mu^{t+1}, \Sigma^t, \alpha^t, c^t, q^t\}$

% \State \textbf{3. Future Latent Prediction:}
% \State $\hat{B}_{t+1} \leftarrow \text{Rasterize}(\hat{G}_{t+1})$ \Comment{Project to BEV latent}

% % \State \textbf{4. Ground Truth Generation (Frozen Teacher):}
% % \State $G_{t+1}^{GT} \leftarrow \mathcal{F}_{perc}(I_{t+1})$
% % \State $B_{t+1} \leftarrow \text{Rasterize}(G_{t+1}^{GT})$

% % \State \textbf{5. Optimization:}
% % \State $\mathcal{L}_{bev} \leftarrow || \hat{B}_{t+1} - B_{t+1} ||_2$ \Comment{Eq. 6}
% % \State Update $\theta_{flow}$ to minimize $\mathcal{L}_{bev}$
% \end{algorithmic}
% \end{algorithm}

\begin{figure*}[t]
  \centering
  \includegraphics[width=1.8\columnwidth]{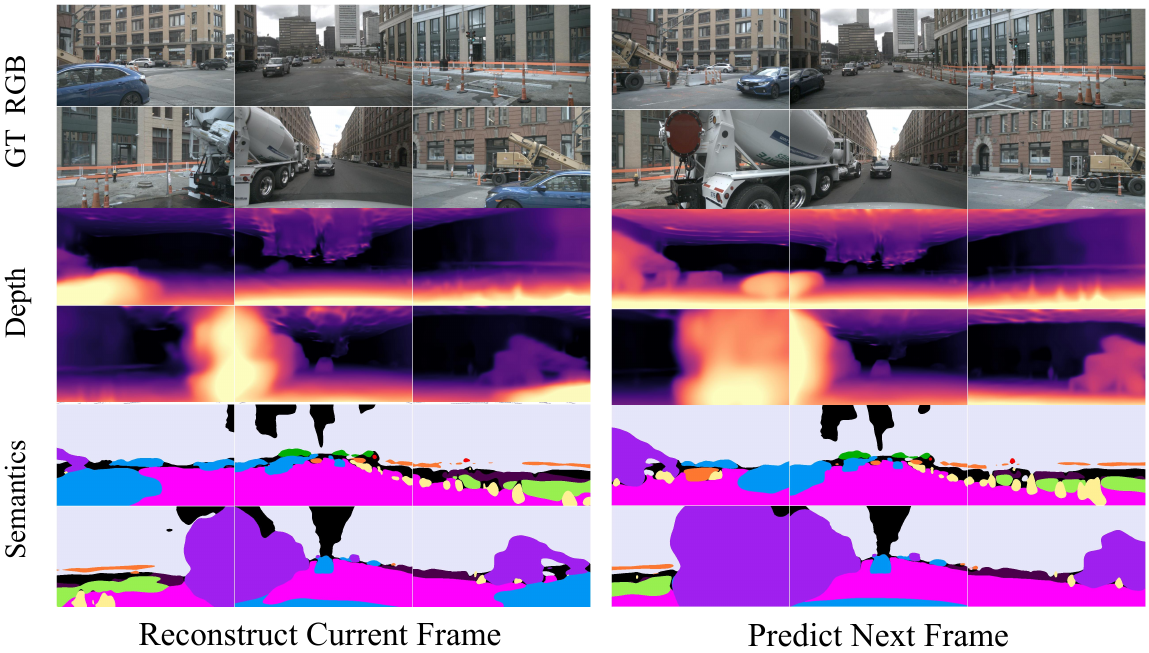}
  \caption{\textbf{Qualitative results of depth and semantic reconstruction and forecasting during the pre-training.} The left column shows the depth and semantic reconstruction at the current frame while the right column illustrates the depth and semantic images rendered by predicted Gaussians at the next frame compared with the GT RGB images.}
  \label{fig:recon}
\end{figure*}

\begin{algorithm}[t]
	\caption{Latent World Model with Gaussian Flow.}          
	\label{alg:flow}                        
	\begin{algorithmic}
		\renewcommand{\algorithmicrequire}{\textbf{Input:}}
		\renewcommand{\algorithmicensure}{\textbf{Output:}}
		% \Require $G_t = \{\mu_k^t, \Sigma_k^t, \alpha_k^t, c_k^t, q_k^t\}$, $T_{ego}^{t \rightarrow t+1}$ 
		\Require $G_t = \{\mu^t, \Sigma^t, \alpha^t, c^t, q^t\}$, $T_{ego}^{t \rightarrow t+1}$ 
		\Ensure $\hat{B}_{t+1}$
		\State \textit{\# predict the absolute flow for each Gaussian}
        \State $\Delta \mu^t \leftarrow \textbf{FlowHead}(q^t)$
		\State \textit{\# apply flow and align ego motion}
        % \For{each Gaussian $k$}
        %     \State $\mu_k^{t+1} \leftarrow T_{ego}^{t \rightarrow t+1}(\mu_k^t + \Delta \mu_k^t)$ 
        % \EndFor
        \State $\mu^{t+1} \leftarrow T_{ego}^{t \rightarrow t+1}(\mu^t + \Delta \mu^t)$ 
		\State \textit{\# update Gaussian set}
        \State $\hat{G}_{t+1} \leftarrow \{\mu^{t+1}, \Sigma^t, \alpha^t, c^t, q^t\}$
        \State \textit{\# predict latent BEV by rasterization}
        \State $\hat{B}_{t+1} \leftarrow \textbf{Rasterize}(\hat{G}_{t+1})$

		\State $\hat{B}_{t+1}$
	\end{algorithmic}
\end{algorithm}

\begin{algorithm}[t]
	\caption{Latent World Model with Ego Planning.}          
	\label{alg:plan}                        
	\begin{algorithmic}
		\renewcommand{\algorithmicrequire}{\textbf{Input:}}
		\renewcommand{\algorithmicensure}{\textbf{Output:}}
		\Require $G_t$, $Q_{wp}$    
		\Ensure  $\hat{B}_{t+1}$
        \State \textit{\# obtain current latent BEV by rasterization}
        \State $B_t \leftarrow \textbf{Rasterize}({G}_t)$
		\State \textit{\# extract scene queries from BEV}
        \State $Q_{scene}^t \leftarrow \textbf{ExtractQueries}(B_t)$
        \State $Q_{scene}^t \leftarrow \textbf{SelfAttn}(Q_{scene}^t)$
		\State \textit{\# predict ego trajectory}
        \State $Q_{wp}' \leftarrow \textbf{CrossAttn}(Q_{wp}, Q_{scene}^t)$ 
        \State $\hat{T} \leftarrow \textbf{MLP}(Q_{wp}')$ 
        \State \textit{\# motion-aware layer norm conditioned on $\hat{T}$}
        \State $Q_{norm} \leftarrow \textbf{MLN}(Q_{scene}^t, \hat{T})$
        \State \textit{\# predict next scene queries}
        \State $Q_{scene}^{t+1} \leftarrow \textbf{SelfAttn}(Q_{norm})$
        \State \textit{\# reconstruct future BEV}
        \State $\hat{B}_{t+1} \leftarrow \textbf{Fuse}(Q_{scene}^{t+1}, B_t)$
        \State $\hat{B}_{t+1}$
	\end{algorithmic}
\end{algorithm}

\myparagraph{Details of Latent World Model with Ego Planning.}
Regarding the latent world model guided by ego planning, we list the pseudo code step by step in Alg. \ref{alg:plan}. Our target is to predict the next latent BEV $\hat{B}_{t+1}$ based on the current Gaussian $G_t$ coupled with ego planning. Firstly, we rasterize current Gaussian $G_t$ to latent BEV ${B}_{t}$~\cite{lu2025toward}. To improve computational efficiency, we extract scene queries $Q_{scene}^t$ from the current BEV ${B}_{t}$ motivated by~\cite{li2024navigation}. The following self-attention module is used to capture global context. To predict ego trajectory, the randomly initialized waypoint queries $Q_{wp}$ interact with $Q_{scene}^t$ through cross attention, followed by a MLP head for predicting ego trajectory $\hat{T}$. 
Then we apply a Motion-Aware Layer Normalization (MLN)~\cite{wang2023exploring} layer for conditioning scene queries $Q_{scene}^t$ on the predicted trajectory $\hat{T}$. The obtained $Q_{norm}$ predicts the next scene queries 
$Q_{scene}^{t+1}$ by self attention. Finally, we predict future BEV $\hat{B}_{t+1}$ by fusing $Q_{scene}^{t+1}$ with the current BEV ${B}_{t}$ following~\cite{li2024navigation}.

% \begin{algorithm}[t]
% \caption{Latent World Model with Ego Planning}
% \label{alg:motion_planning}
% \begin{algorithmic}[1]
% \Require 
%     $B_t$, $Q_{wp}$
% \Ensure $\hat{B}_{t+1}$

% \State \textbf{1. Context Extraction:}
% \State $Q_{scene}^t \leftarrow \text{ExtractQueries}(B_t)$ \Comment{Extract scene queries from BEV}
% \State $Q_{scene}^t \leftarrow \text{SelfAttn}(Q_{scene}^t)$

% \State \textbf{2. Motion Planning:}
% \State $Q_{wp}' \leftarrow \text{CrossAttn}(Q_{wp}, Q_{scene}^t)$ \Comment{Interaction}
% \State $\hat{T} \leftarrow \text{MLP}(Q_{wp}')$ \Comment{Predict trajectory}
% \State $\mathcal{L}_{reg} \leftarrow || \hat{T} - T ||_1$ \Comment{Eq. 7: Imitation loss}

% \State \textbf{3. Future Latent Prediction (Conditioned on $\hat{T}$):}
% \State $Q_{norm} \leftarrow \text{MLN}(Q_{scene}^t, \hat{T})$ \Comment{Motion-Aware Layer Norm}
% \State $Q_{scene}^{t+1} \leftarrow \text{SelfAttn}(Q_{norm})$ \Comment{Predict next scene queries}
% \State $\hat{B}_{t+1} \leftarrow \text{Fuse}(Q_{scene}^{t+1}, B_t)$ \Comment{Reconstruct future BEV}

% % \State \textbf{4. Optimization:}
% % \State $\mathcal{L}_{bev} \leftarrow || \hat{B}_{t+1} - B_{t+1} ||_2$ \Comment{Eq. 6}
% % \State $\mathcal{L}_{plan} \leftarrow \mathcal{L}_{reg} + \mathcal{L}_{bev}$ \Comment{Eq. 8: Total loss}
% % \State Update model to minimize $\mathcal{L}_{plan}$
% \end{algorithmic}
% \end{algorithm}

% \myparagraph{Details of Constructing Latent BEV.}

\begin{figure*}[t]
  \centering
  \includegraphics[width=1.8\columnwidth]{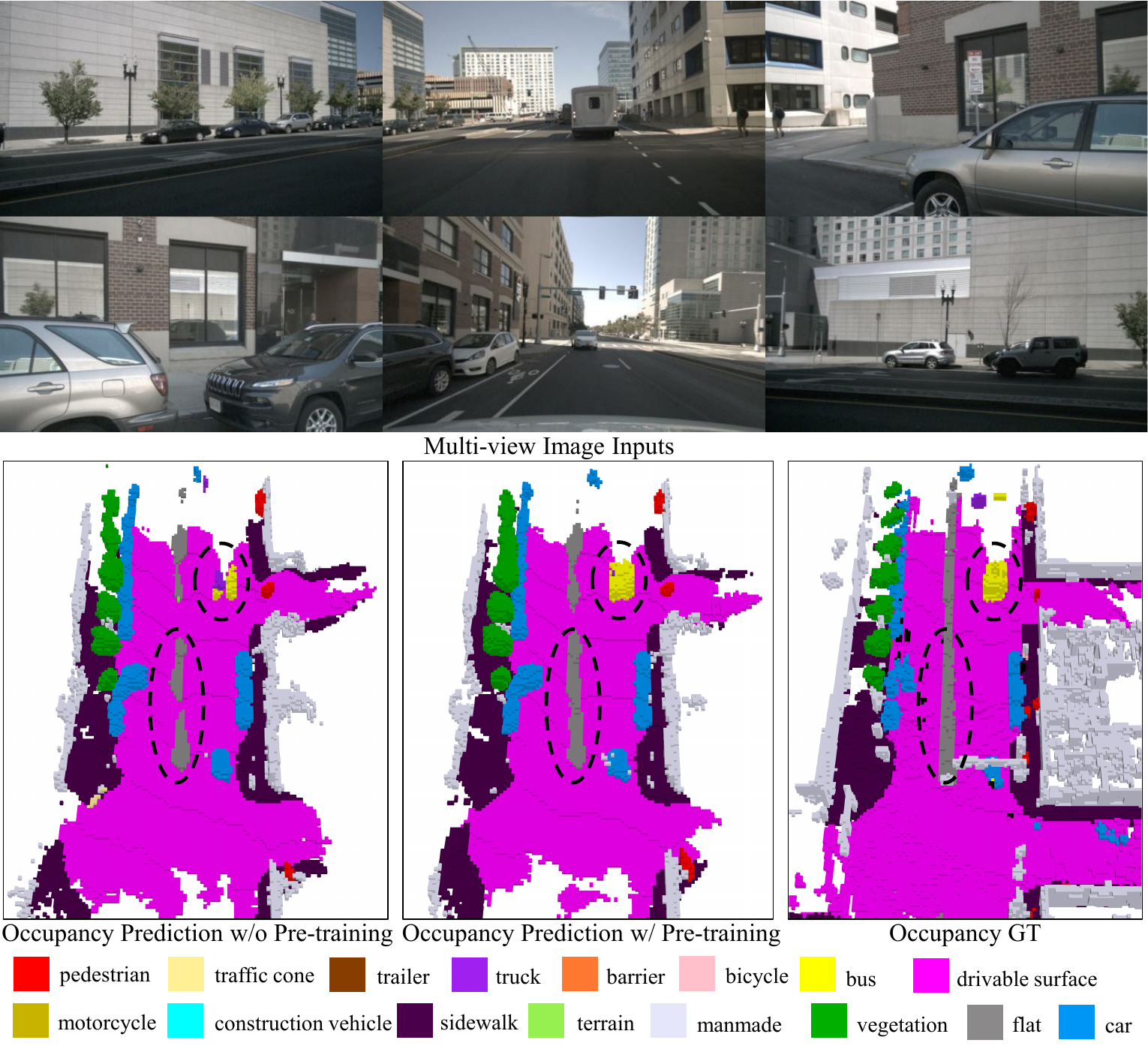}
  \caption{\textbf{Qualitative comparison of 3D Occupancy Perception over GT, baseline without pre-training and ours with pre-training.}}
  \label{fig:3docc}
\end{figure*}

\begin{table}[h]
\caption{\textbf{Ablation study on ``Dual'' and ``Unified''.} }
\centering
\footnotesize
\setlength{\tabcolsep}{4mm}
\renewcommand\arraystretch{1.15}
\begin{tabular}{lccc}
\toprule
Model & mIoU (\%) $\uparrow$ & L2 (m) $\downarrow$ & Col. (\%) $\downarrow$ \\
\midrule
Unified & 18.9 & 0.58 & 0.22 \\
Dual & \textbf{19.3} & \textbf{0.46} & \textbf{0.19} \\
\bottomrule
\end{tabular}
\label{tab:supp4}
\end{table}

\begin{table}[h]
\caption{\textbf{Ablation study on the number of 3D Gaussians.} }
\centering
\footnotesize
\setlength{\tabcolsep}{4mm}
\renewcommand\arraystretch{1.15}
\begin{tabular}{lcc}
\toprule
Nums & mIoU (\%) $\uparrow$ & IoU (\%) $\uparrow$ \\
\midrule
7500 & 18.35 & 28.83 \\
25600 & 19.30 & \textbf{30.56} \\
51200 & \textbf{19.43} & 29.79 \\
\bottomrule
\end{tabular}
\label{tab:supp1}
\end{table}

\begin{table}[h]
\caption{\textbf{Ablation study on the number of future frames.} }
\centering
\footnotesize
\setlength{\tabcolsep}{4mm}
\renewcommand\arraystretch{1.15}
\begin{tabular}{lcc}
\toprule
Nums & mIoU (\%) $\uparrow$ & IoU (\%) $\uparrow$ \\
\midrule
0 & 18.83 & 30.12 \\
1 & \textbf{19.30} & \textbf{30.56} \\
2 & 19.23 & 30.52 \\
4 & 19.06 & 29.20 \\
\bottomrule
\end{tabular}
\label{tab:supp2}
\end{table}

\begin{table}[h]
\caption{\textbf{The computation cost analysis of all three tasks.} }
\centering
\footnotesize
\setlength{\tabcolsep}{4mm}
\renewcommand\arraystretch{1.15}
\begin{tabular}{lcc}
\toprule
Task & Memory & Latency \\
\midrule
Perception & 4.4GB & 327ms \\
Forecasting & 6.8GB & 684ms \\
Planning & 4.2GB & 274ms \\
\bottomrule
\end{tabular}
\label{tab:supp3}
\end{table}

\section{Additional Experimental Results}
% Ablation on the number of queries and Gaussians. FPS is measured on an A100 GPU.

\myparagraph{Comparing Dual Models with Unified Model.} The motivation of using two world models is that
decoupling the training of Gaussian flow and ego planning is imperative for reducing learning complexity.
% As illustrated in Eq. \ref{eq:flowpred}, the predicted Gaussian mean is propagated through a combination of Gaussian flow and ego-motion alignment. For the Gaussian-flow-guided model, leveraging GT ego motion reduces the learning complexity of Gaussian flow. Conversely, explicitly integrating Gaussian flow learning into the motion planning model would increase the latter’s learning burden. This is because self-supervised Gaussian flow lacking explicit supervision introduces unnecessary complexity to the planning, since the two items are coupled as shown in Eq. \ref{eq:flowpred}.

As illustrated in Eq.5 (main paper), the predicted Gaussian mean is propagated through a combination of Gaussian flow and ego-motion alignment. For the Gaussian-flow-guided model, leveraging GT ego motion reduces the learning complexity of Gaussian flow. Conversely, explicitly integrating Gaussian flow learning into the motion planning model would increase the latter’s learning burden. This is because self-supervised Gaussian flow lacking explicit supervision introduces unnecessary complexity to the planning, since the two items are coupled as shown in Eq.5 (main paper).

To verify this, we supplement an ablation experiment comparing the ``Dual" models with the ``Unified" model, where the predicted ego motion is utilized for motion transformation in place of GT in Eq. \ref{eq:flowpred} for unifying learning of both flow and planning. The results in Tab. \ref{tab:supp4} confirm that the decoupled design achieves superior performance in perception and planning, justifying its necessity. 

\myparagraph{Number of Gaussians.} We conduct an ablation study (see Tab.~\ref{tab:supp1}) to evaluate the influence of the number of 3D Gaussian queries on the 3D occupancy perception performance. Testing configurations of 7,500, 25,600, and 51,200 Gaussians, the results reveal a trade-off: The mIoU (19.43) peaks at 51,200 Gaussians, indicating superior semantic granularity and expressive power. The IoU (30.56) achieves its best geometric accuracy with 25,600 Gaussians, suggesting this configuration strikes the optimal balance between density and reconstruction efficiency. We prioritize the configuration with 25,600 Gaussians for our final model.

\myparagraph{Number of Future Frames.} 
We investigate the influence of the predicted number of future frames for pre-training stage 2 regarding Gaussian-flow-guided latent world model. 3D occupancy perception is selected for an ablation study. As shown in Tab.~\ref{tab:supp2}, we predict future Gaussians with four settings including ``0'', ``1'', ``2'', and ``4'' future frames. The setting ``0'' denotes that we only reconstruct images without future prediction. By incorporating the prediction of the future frame $N=1$, the performance peaks at $19.30$ mIoU and $30.56$ IoU, demonstrating  the beneficial impact of the Gaussian-flow-guided latent world model. The inclusion of additional future frames may slightly impair performance, potentially attributable to the greater scene changes caused by more distant future frames, which complicates accurate future prediction.

\myparagraph{Inference Latency and Memory Analysis.} The dual world models are only used during the pre-training to facilitate representation learning in Gaussian perception module. For the inference stage, only the perception module and each task-specific head are used. Thus, the pre-training does not introduce additional inference latency. We measure inference latency and memory for each task on one 
% RTX 3090 
GPU with batch size one. The evaluated results in Tab. \ref{tab:supp3} show lightweight memory and latency cost.

\section{BEV Supervision Discussion}
\noindent BEV rasterization is used only for temporal supervision, not as the representation itself. Crucially: 1. All geometric reasoning happens in 3D Gaussian space during both pre-training and inference. 2. BEV features are derived from 3D Gaussians via differentiable rasterization that preserves height information through vertical stacking (following GaussianLSS~\cite{lu2025toward}).

\section{Visualization Results}

\myparagraph{Depth and Semantic Reconstruction and Forecasting.} To validate the effectiveness of our holistic Gaussian-centric pre-training, we present qualitative visualization results for depth and semantic image rendering in Fig.~\ref{fig:recon}. During the pre-training stage, the model is tasked with reconstructing and forecasting the 3D scene geometry and semantics under self-supervision. The rendered depth maps (shown in the middle rows) exhibit sharp boundaries and accurate depth stratification, showing the ability for recovering detailed geometric structures. The rendered semantic maps (shown in the bottom rows) demonstrate high consistency with the semantic priors. Based on the excellent reconstruction performance in the left column of Fig. \ref{fig:recon}, the model exhibits accurate depth and semantic prediction at the next frame in the right column, meaning that the pre-training paradigm is able to model the temporal evolution for the driving scene.

\myparagraph{3D Occupancy Perception.} We provide additional visualization results for the 3D occupancy perception task in Fig.~\ref{fig:3docc}. DLWM significantly outperforms the baseline without pre-training in both geometric completeness and semantic consistency. While the baseline suffers from fragmented structures and boundary misclassifications due to sparse observations, DLWM generates significantly more complete geometric structures and more accurate semantic classifications (e.g., for vehicles and vegetation) compared to the baseline.

\begin{figure*}[t]
  \centering
  \includegraphics[width=2\columnwidth]{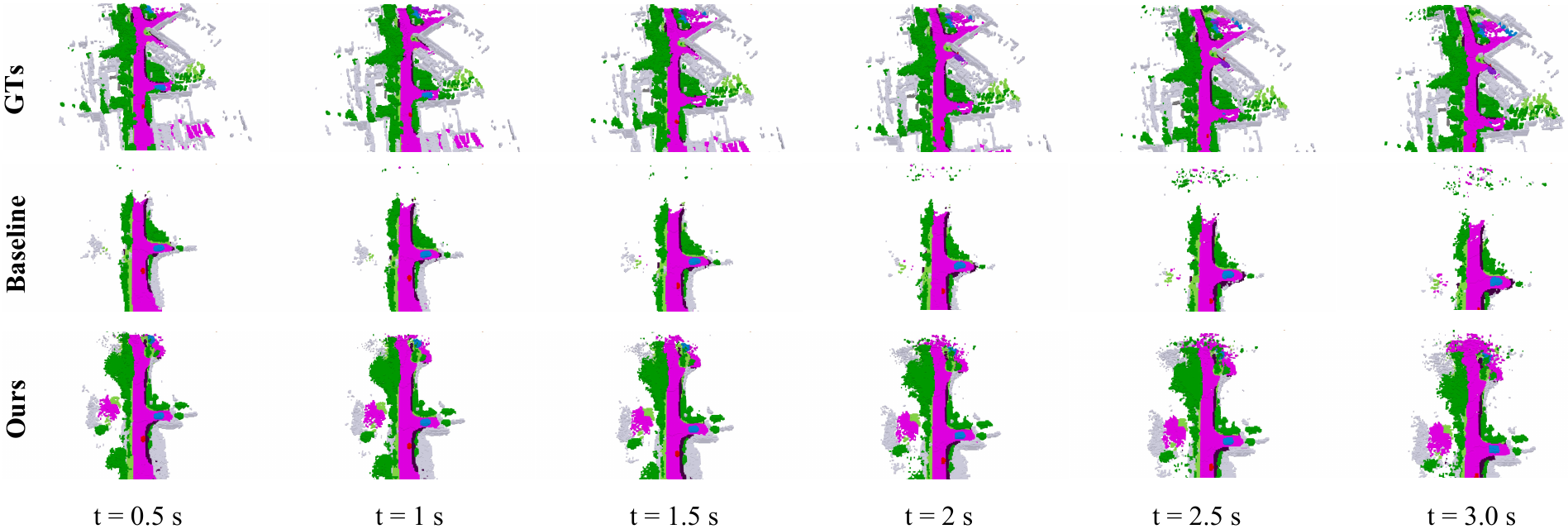}
  \caption{\textbf{Qualitative comparison of 4D Occupancy Forecasting over GTs, baseline without pre-training and ours with pre-training.} }
  \label{fig:4docc}
\end{figure*}

\myparagraph{4D Occupancy Forecasting.} We display the forecasted occupancy scenes at future time steps (e.g., 1s, 2s, and 3s) in Fig.~\ref{fig:4docc}. Compared to the baseline without pre-training, DLWM generates temporally consistent predictions with sharper geometric details. In contrast to the baseline, which produces blurred or static predictions, DLWM is pre-trained with Gaussian-flow-guided latent world model to explicitly predict Gaussian displacement, ensuring accurate trajectory forecasting for dynamic agents.

% \myparagraph{Motion Planning.}

\end{document}